\pdfoutput=1
\documentclass[11pt]{article}

\usepackage[preprint]{acl}

\usepackage{times}
\usepackage{booktabs}
\usepackage{latexsym}
\usepackage{multirow}
\usepackage{arydshln}  % 如果你使用的是这个包提供的 \hdashline
\usepackage{amssymb}
\usepackage{amsmath}
\usepackage{makecell}  % 加载 makecell 包来支持 \makecell
\usepackage{url}
\usepackage{graphicx}
\usepackage{microtype}
\usepackage{multicol}
\usepackage{xcolor}
\usepackage{lipsum}
\usepackage{enumitem}
\usepackage{pifont}
\usepackage{colortbl}
\usepackage{tabularx}
\usepackage{caption} 
\usepackage{float}
\usepackage{amsmath,amsfonts,amssymb,bbm}
\usepackage{adjustbox}
\usepackage{fvextra}
\usepackage[breakable]{tcolorbox}
\usepackage{subcaption}
\usepackage{hyperref}
\usepackage{amsmath}
\usepackage{amssymb}
\usepackage{xltabular}
\usepackage{booktabs}
\usepackage{tcolorbox}
\usepackage{listings}
\newcommand{\piref}{\pi_\text{ref}}

\usepackage[T1]{fontenc}

\usepackage[utf8]{inputenc}

\usepackage{microtype}

\usepackage{inconsolata}

\definecolor{green}{RGB}{36, 214, 36}
\definecolor{red}{RGB}{235, 30, 30}

\usepackage[utf8]{inputenc}
\usepackage[T1]{fontenc}

\usepackage{CJKutf8}

\usepackage{CJKutf8}
\usepackage{booktabs} % 用于更好的表格线条

\usepackage{CJK}
\usepackage{xcolor}
\usepackage{ulem}

\usepackage{xspace}
\newcommand{\method}{R-PRM\xspace}
\newcommand\blfootnote[1]{%
\begingroup
\renewcommand\thefootnote{}\footnote{#1}%
\addtocounter{footnote}{-1}%
\endgroup
}

\renewcommand{\paragraph}[1]{\noindent\textbf{#1}}

\title{R-PRM: Reasoning-Driven Process Reward Modeling}

\author{
    Shuaijie She$^{1}$$^*$\text{,} \textbf{Junxiao Liu}$^{1}$$^*$\text{,} \textbf{Yifeng Liu}$^{1}$\text{,} \textbf{Jiajun Chen}$^{1}$\text{,} \textbf{Xin Huang}$^{2}$\textbf{,} \textbf{Shujian Huang}$^{1}$$^\dagger$   \\
    $^{1}$ \text{National Key Laboratory for Novel Software Technology, Nanjing University} \\
    $^{2}$ \text{China Mobile Communications Company Limited Research Institute} \\
    \small\texttt{\{shesj,junxiaoliu,yfliu\}@smail.nju.edu.cn}, 
    \small\texttt{chenjj@nju.edu.cn}, \\
    \small\texttt{huangxinyjy@chinamobile.com},
    \small\texttt{huangsj@nju.edu.cn} \\
}

\begin{document}
\maketitle
\blfootnote{$^*$ Equal contributions.}
\blfootnote{$^\dagger$Corresponding author.}
\begin{abstract}
Large language models (LLMs) inevitably make mistakes when performing step-by-step mathematical reasoning. Process Reward Models (PRMs) have emerged as a promising solution by evaluating each reasoning step. However, existing PRMs typically output evaluation scores directly, limiting both learning efficiency and evaluation accuracy, which is further exacerbated by the scarcity of annotated data. To address these issues, we propose Reasoning-Driven Process Reward Modeling (\textbf{R-PRM}). First, we leverage stronger LLMs to generate seed data from limited annotations, effectively bootstrapping our model's reasoning capabilities and enabling comprehensive step-by-step evaluation. Second, we further enhance performance through preference optimization, without requiring additional annotated data. Third, we introduce inference-time scaling to fully harness the model's reasoning potential. Extensive experiments demonstrate R-PRM's effectiveness: on ProcessBench and PRMBench, it surpasses strong baselines by 11.9 and 8.5 points in F1 scores, respectively. When applied to guide mathematical reasoning, R-PRM achieves consistent accuracy improvements of over 8.5 points across six challenging datasets. Further analysis reveals that R-PRM exhibits more comprehensive evaluation and stronger generalization capabilities, thereby highlighting its significant potential~\footnote{The project is available at \url{https://github.com/NJUNLP/R-PRM}}.
\end{abstract}

\section{Introduction}
Recently, large language models (LLMs) have demonstrated significant progress in solving challenging mathematical problems through chain-of-thought reasoning~\cite{cot,qwne25math,DeepSeekMath}. However, LLMs still tend to make reasoning errors, undermining the reliability of their solutions and hindering their ability to arrive at correct answers.

Therefore, Process Reward Model (PRM) has been proposed to further improve models' reasoning ability~\cite{verify-step-by-step}. Unlike Outcome Reward Models (ORM) that only focus on the final results, PRMs evaluate each reasoning step in a more fine-grained manner and achieve better performance and generalization, such as identifying and mitigating process errors, while also exhibiting stronger generalization capabilities~\cite{verify-step-by-step,Math-shepherd}. 

A primary challenge in PRM development arises from data scarcity. While human annotation can provide high-quality process-level labels~\cite{verify-step-by-step}, it incurs substantial costs. Alternative automated approaches, such as Monte Carlo (MC) methods that estimate step correctness based on the probability of reaching the correct final answer~\cite{Math-shepherd,openr,auto-process-supervision}, or methods that use stronger language models as judges for data filtering~\cite{qwenprm}, have shown some promise. However, these methods either require significant computational resources or still struggle with noise and bias, leaving the challenge of sufficient high-quality training data unresolved. 

Moreover, existing process reward models directly provide evaluations based on the given steps.  We argue that for challenging process-level evaluation tasks, this direct evaluation approach constrains the model's learning process and reduces learning efficiency. Moreover, it lacks interpretability, as it fails to identify why specific steps are incorrect, making it difficult to provide constructive feedback for improvement.

To address these issues, we propose a Reasoning-Driven Process Reward Modeling (\textbf{R-PRM}) framework, which utilizes reasoning over each intermediate step to improve process-level evaluation. The framework consists of three key components: First, we construct seed data by prompting stronger LLMs with a small amount of human-annotated step-level labels and subsequently fine-tune Qwen2.5-Math-7B-Instruct. Through this reasoning-centric paradigm, our model develops the capability to perform comprehensive and transparent analyses for evaluating complex solution steps of challenging questions. Second, the generative evaluation paradigm enables us to apply preference optimization, improving model capabilities by encouraging the generation of evaluation processes that lead to correct judgments, without requiring additional data generation. Finally, we exploit the generative nature of our evaluation paradigm at inference time, allowing multiple evaluation processes to be sampled for a more comprehensive and robust assessment perspective.

When evaluated on ProcessBench and PRMBench, our \method achieves F1 score improvements of 11.9 and 8.5 points, respectively, over the strongest baseline trained on the same data. Furthermore, when used to guide policy model reasoning via Best-of-N and Guided Search strategies, our approach improves accuracy by average margins of 8.6 and 8.4 points over the Pass@1 baseline across six challenging math datasets, outperforming both majority voting and all existing PRM baselines. Further analysis reveals our three key additional advantages: (1) comprehensive evaluation coverage through multi-dimension analysis, (2) enhanced generalization capability across diverse datasets, and (3) progressive accuracy improvement with increased reasoning budgets, suggesting significant potential for practical reasoning-system optimization.

\section{Related Work}
\subsection{Mathematical Reasoning}
Recent studies have demonstrated that LLMs exhibit enhanced reasoning capabilities when generating step-by-step solutions before providing the final answers~\cite{cot}. Building on this insight, several pioneering works have focused on developing large-scale mathematical datasets with high-quality reasoning annotations for fine-tuning of LLMs~\cite{WizardMath,selfinstruct,DeepSeekMath,qwne25math}. However, even when models arrive at correct final answers, their intermediate reasoning steps may contain critical errors. This discrepancy undermines the reliability of their problem-solving processes and poses significant obstacles for future model improvements~\cite{processbench}.

Parallel advancements~\cite{scaling-test-time,o1,deepseekr1,qwq} in inference-time have demonstrated that increasing the computational budget to enable multiple reasoning attempts, coupled with majority voting mechanisms for answer selection, can achieve remarkable accuracy improvements.
% Furthermore, at the inference stage, researchers have discovered that increasing the computational budget to enable multiple reasoning attempts, combined with majority voting for final answer determination, can substantially improve accuracy rates~\cite{scaling-test-time,o1,deepseekr1,qwq}. This approach, despite its increased computational cost, has proven particularly effective in enhancing the overall performance on mathematical reasoning tasks.

% At the same time, at the inference stage, researchers have discovered that increasing computational budget to enable multiple reasoning attempts, combined with majority voting to determine the final answer, can substantially improve accuracy rates~\cite{scaling-test-time,o1,deepseekr1,qwq}. This approach, despite its higher computational cost, has proven particularly effective in enhancing the overall performance of mathematical reasoning tasks.

% To address these limitations, researchers have explored inference-time optimization strategies. Specifically, increasing computational budgets to enable multiple reasoning attempts, coupled with majority voting mechanisms for answer selection, has demonstrated substantial improvements in model accuracy~\cite{scaling-test-time,o1,deepseekr1,qwq}. While this approach incurs higher computational costs, it has emerged as a particularly effective technique for enhancing mathematical reasoning performance.

\subsection{Reward Modeling of Reasoning}
Reward models are introduced to further improve mathematical reasoning by enhancing training data quality, guiding model learning~\cite{verify-step-by-step,first-orm,first-prm}, and guiding the policy model's reasoning process through Best-of-N and Guided-Search methods~\cite{Math-shepherd,qwenprm}.

Currently, reward models are typically categorized into Outcome Reward Models (ORM) and Process Reward Models (PRM)~\cite{verify-step-by-step}. ORM focuses on providing an overall evaluation based on whether the correct answer is ultimately obtained~\cite{first-orm}.. In contrast, PRM provides a fine-grained evaluation for each reasoning step, and many works have shown that PRM can achieve better results~\cite{verify-step-by-step,first-prm}. However, data for PRM is extremely scarce, and the annotation  is costly~\cite{verify-step-by-step,Math-shepherd,auto-process-supervision}. Some studies explore automatic synthesis strategies, such as using Monte Carlo (MC) estimation methods~\cite{Math-shepherd,auto-process-supervision}. However, MC methods incur a large computational cost and inevitably introduce bias and noise~\cite{processbench}. \cite{qwenprm} propose combining MC with LLM as a judge, helping to reduce noise. However, the quality and quantity of step-level reasoning evaluation data are still limited, and this remains an unsolved challenge.

\section{Method}

\begin{figure*}[ht]
\centering
\setlength{\topskip}{0pt}  % 减少顶部间距
\setlength{\belowcaptionskip}{0pt}   % 减少caption下方间距
\includegraphics[scale=0.60]{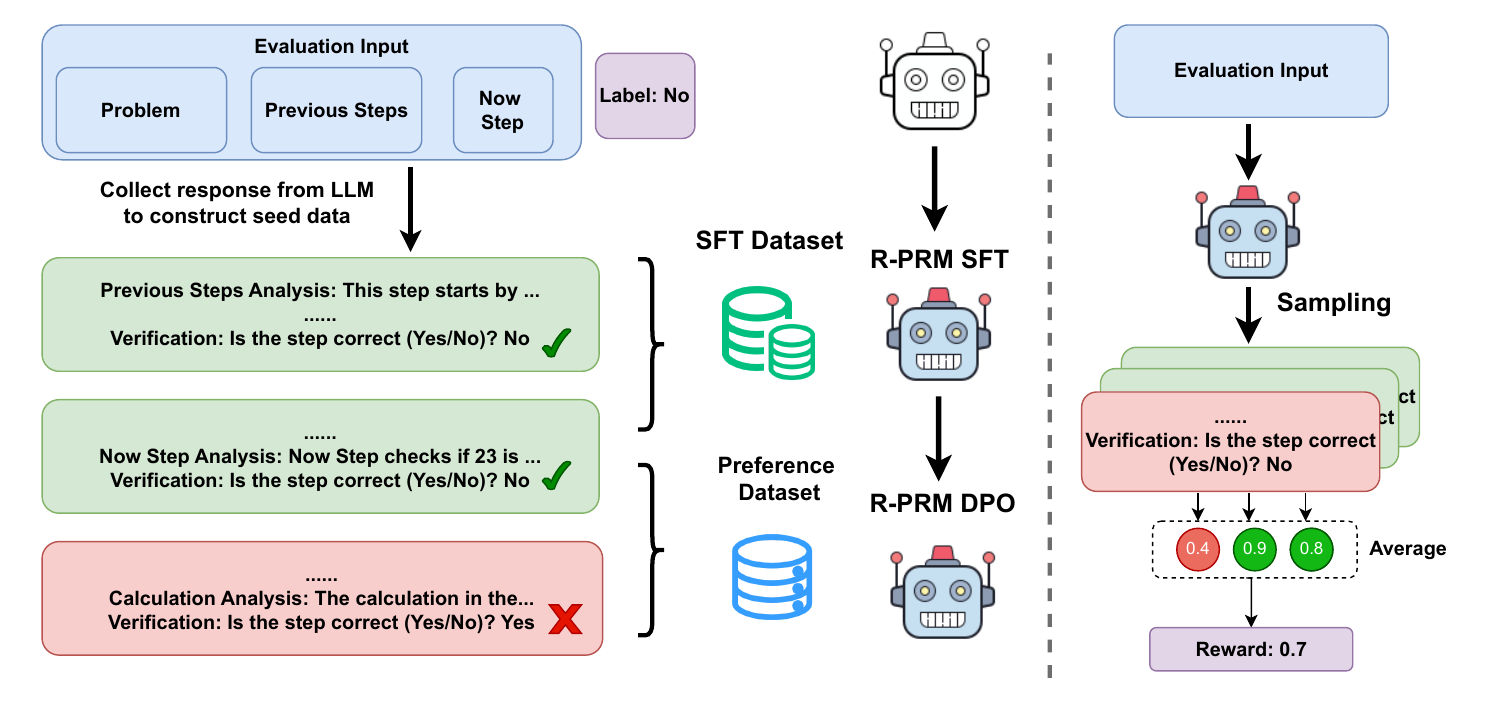}
\caption{Illustration of our framework. For brevity, only partial analytical reasoning trajectories are shown. White robots indicate initial models, while colored ones represent models after our training procedure.
}
\label{method-fig}
\vspace{-\baselineskip} % 可选：进一步压缩下方间距
\end{figure*}
\subsection{Reasoning for Process Reward Modeling}
Given a mathematical problem $Q$, the policy model generates a sequential chain-of-reasoning process $S = \{s_1, s_2, ..., s_n\}$, where each reasoning step $s_i$ is generated conditioned on both $Q$ and the preceding steps $\{s_1, ..., s_{i-1}\}$. To evaluate the quality of each reasoning step, current process-level reward models employ a direct prediction mechanism that assigns a score to each step. This evaluation process can be formally expressed as:
\begin{equation*}
    R_i = M(Q, s_1, ..., s_i)
\end{equation*}
where $M(\cdot)$ represents the reward model that outputs a scalar reward $R_i$. 

Unlike previous approaches, we propose a reasoning-driven process-level reward model $G$ that operates in a single generation process to perform two phases. First, $G$ generates a comprehensive analysis $A_i$ of each reasoning step $s_i$, consisting of multiple analytical dimensions: examining historical reasoning steps, assessing the objective and data sources of the current step, verifying its coherence with preceding steps, and validating the computational transformations involved. Then, following the analysis,  model generates a judgment $J_i$ which is is a natural language phrase indicating the correctness of the step, expressed as "yes" or "no":
\begin{equation*}  
A_i = G(Q, s_1, ..., s_i)  
\end{equation*}  
\begin{equation*}  
J_i = G(Q, s_1, ..., s_i, A_i)  
\end{equation*}  

% For \textit{hard labels}, these strings are directly mapped to binary scores (0 or 1), while for \textit{soft labels}, the probability of "yes" is adopted as the score of current step. 

We prompt a stronger LLM with samples from PRM800K to generate ($Q$, $s_{1:i}$, $A_i$, $J_i$) tuples~\footnote{The prompt we used is listed in Appendix \ref{sec:construction-data-prompt}}. We retain only those evaluation analyses that produce a judgment consistent with the human label. Subsequently, we concatenate the analysis and score as the target sequence. Let $Y_i$ denote the complete output sequence for $s_i$:
\begin{equation*}
Y_i= A_i \oplus J_i = \{y_{1}, y_{2}, ..., y_{t}\}
\end{equation*}
\begin{equation*}
\mathcal{L}_{\text{SFT}} = -\sum_{j=1}^{t} \log p(y_{j}|Q,s_{1:i},y_{1:j-1})
\end{equation*}
where $y_j$ denotes the $j$-th token in the output sequence $Y_i$, and $t$ is the total length of the sequence. This is equivalent to standard instruction tuning, where the model learns to generate both the analysis and the judgment in a single forward pass.

\vspace{-0.5em}
\subsection{Process Reward Modeling Meta-Optimization}

Through supervised fine-tuning (SFT), our model learns to effectively evaluate complex reasoning trajectories in mathematical problems by leveraging its reasoning capabilities. We further enhance the model’s performance by optimizing its reasoning process. More specifically, we leverage preference optimization methods like Direct Preference Optimization (DPO)~\cite{dpo} to encourage the model to generate reasoning processes that yield correct judgments.

DPO involves an input pair $(Y^{w}, Y^{l})$, where $Y^{w}$ is favored over $Y^{l}$. Correspondingly, we sample multiple evaluation reasoning processes, which are categorized into two groups based on whether their final predictions match the annotated labels. We treat consistent trajectories as $Y^{w}$ and inconsistent ones as $Y^{l}$ to construct preference pairs. We freeze the original supervised fine-tuned LLM as the reference policy $\pi_\text{ref}$ and optimize our model using the following loss function:
\begin{flalign*}
    &\mathcal{L}_\text{DPO}(\pi_{\theta}; \piref) =  -\mathbb{E}_{(x, Y^{w}, Y^{l})\sim \mathcal{D}} & \nonumber \\
&\left[\log \sigma \left(\beta \log \frac{\pi_{\theta}(Y^{w}\mid x)}{\piref(Y^{w}\mid x)} - \beta \log \frac{\pi_{\theta}(Y^{l}\mid x)}{\piref(Y^{l}\mid x)}\right)\right]
\label{eq:our-dpo}
\end{flalign*}

\subsection{Inference Time Scaling Strategy}

% Leveraging our PRM's capability to generate diverse evaluation processes, we could adopt a scalable inference strategy. Specifically, during the inference phase, we allocate additional computational budget to enable multiple reasoning trajectory generations. These trajectories are then systematically aggregated through majority voting mechanisms to derive final predictions. This approach effectively mitigates potential reasoning inconsistencies and stochastic variations, thereby enhancing both the robustness and accuracy of our evaluation pipeline. 

% In \textit{hard-label} settings, we simply use the frequency of "yes" judgments as the final score. For \textit{soft-label} settings, we first compute the softmax probabilities between "yes" and "no" responses, then average the probabilities across multiple reasoning trajectories to obtain the final reward:
% \begin{equation*}
% R_i = \frac{1}{K}\sum_{k=1}^K \text{softmax}(\text{R}_i^{(k)})_{\text{yes}}
% \end{equation*}

% where $K$ denotes the number of sampled evaluations. The final judgment is determined by comparing the aggregated rewards:
% \begin{equation*}
% \text{Label} = 
% \begin{cases}
% \text{yes}, & \text{if } R_i > 0.5 \\
% \text{no}, & \text{otherwise}
% \end{cases}
% \end{equation*}

Leveraging our PRM's capability to generate analytical reasoning processes, we propose a scalable inference strategy that enhances evaluation robustness through multiple reasoning trajectories. During inference, for each reasoning step $s_i$, we sample $K$ independent analytical processes as follows:
\begin{equation*}
(A_i^{(k)}, J_i^{(k)}) = G(Q, s_1, ..., s_i), k \in [1,K]
\end{equation*}
where each $A^{(k)}$ represents a distinct analytical reasoning process and $R^{(k)}$ is the corresponding judgment ("yes" or "no"). This multi-trajectory approach helps mitigate potential reasoning inconsistencies and stochastic variations inherent in large language models. To aggregate multiple evaluations, we calculate the average probability of "yes" judgments (using softmax with "no" judgments) as the reward:
\begin{equation*}
R_i = \frac{1}{K}\sum_{k=1}^K P(J_i^{(k)} = \text{"yes"}|Q, s_1, ..., s_i, A_i^{(k)}).
\end{equation*}
% The binary decision of correctness is determined by:
% \vspace{-0.5ex}
% \begin{equation*}
% \text{Label} = 
% \begin{cases}
% 1, & \text{if } R_i > 0.5 \\
% 0, & \text{otherwise}
% \end{cases}
% \end{equation*}

% For aggregating multiple evaluations, we propose two schemes based on different label settings: For \textit{hard-label} settings, we employ majority voting across the $K$ judgments:
% \begin{equation*}
% R_i = \frac{1}{K}\sum_{k=1}^K \mathbbm{1}[R_i^{(k)} = \text{"yes"}]
% \end{equation*}

% For \textit{soft-label} settings, 

\section{Experiment}
\begin{table*}[!t]
\centering
\scriptsize
\renewcommand{\arraystretch}{1.2}
\resizebox{\textwidth}{!}{
    \begin{tabular}{lccccccccccccc}
        \toprule[1.5pt]
        \multirow{2}{*}{\textbf{MODEL}} & \multicolumn{3}{c}{\textbf{GSM8K}} & \multicolumn{3}{c}{\textbf{MATH}} & \multicolumn{3}{c}{\textbf{OLYMPIADBENCH}} & \multicolumn{3}{c}{\textbf{OMNIMATH}} & \multirow{2}{*}{\textbf{Avg. F1}} \\
        \cmidrule(lr){2-4} \cmidrule(lr){5-7} \cmidrule(lr){8-10} \cmidrule(lr){11-13}
         & error & correct & \textbf{F1} & error & correct  & \textbf{F1} & error & correct  & \textbf{F1} & error & correct  & \textbf{F1} \\
        \midrule
        \multicolumn{14}{l}{\textbf{\textit{LLM-as-judge, Proprietary language models}}} \\
        GPT-4o* & 70.0 & 91.2 & 79.2 & 54.4 & 76.6 & 63.6 & 45.8 & 58.4 & 51.4 & 45.2 & 65.6 & 53.5 & 61.9 \\
        o1-mini* & 88.9 & 97.9 & \textbf{93.2} & 83.5 & 95.1 & 88.9 & 80.2 & 95.6 & 87.2 & 74.8 & 91.7 & 82.4 & 87.9 \\
        \midrule
        \multicolumn{14}{l}{\textbf{\textit{LLM-as-judge, Open-source language models}}} \\
        Llama-3.3-70B-Instruct* & 72.5 & 96.9 & 82.9 & 43.3 & 83.2 & 59.4 & 31.0 & 94.1 & 46.7 & 28.2 & 90.5 & 43.0 & 58.0\\
        Qwen2.5-Math-72B-Instruct* & 49.8 & 96.9 & 65.8 & 36.0 & 94.3 & 52.1 & 19.5 & 97.3 & 32.5 & 19.0 & 96.3 & 31.7 & 45.5\\
        Qwen2.5-72B-Instruct* & 62.8 & 96.9 & 76.2 & 46.3 & 93.1 & 61.8 & 38.7 & 92.6 & 54.6 & 36.6 & 90.9 & 52.2 & 61.2 \\
        % QwQ-32B-Preview & 81.6 & 95.3 & \textbf{88.0} & 78.1 & 79.3 & \textbf{78.7} & 61.4 & 54.6 & \textbf{57.8} & 55.7 & 68.0 & \textbf{61.3} & \textbf{71.5}\\
        \midrule
        \multicolumn{14}{l}{\textbf{\textit{PRMs}}} \\
        % \noalign{\vspace{4pt}}
        % \hline
        % \noalign{\vspace{4pt}}
        % \hdashline
        % \multicolumn{14}{l}{\textbf{7B+}} \\
        Math-Shepherd-7B* & 32.4 & 91.7 & 47.9 & 18.0 & 82.0 & 29.5 & 15.0 & 71.1 & 24.8 & 14.2 & 73.0 & 23.8 & 31.5\\
        Math-PSA-7B & 48.3 & 88.1 & 62.4 & 29.5 & 72.7 & 41.9 & 20.7 & 65.8 & 31.5 & 15.4 & 68.9 & 25.2 & 40.3 \\
        RLHFlow-Mistral-8B* & 33.8 & \textbf{99.0} & 50.4 & 21.7 & 72.2 & 33.4 & 8.2 & 43.1 & 13.8 & 9.6 & 45.2 & 15.8 & 28.4 \\
        RLHFlow-DeepSeek-8B* & 24.2 & 98.4 & 38.8 & 21.4 & 80.0 & 33.8 & 10.1 & 51.0 & 16.9 & 10.9 & 51.9 & 16.9 & 26.6 \\
        Llemma-PRM800K-7B & 36.7 & 71.0 & 48.4 & 39.2 & 47.8 & 43.1 & 33.1 & 25.1 & 28.5 & 35.4 & 31.5 & 33.4 & 38.4 \\
        %Llemma-MetaMath-7B & 20.8 & 92.2 & 33.9 & 16.7 & 89.2 & 28.1 & 8.0 & 84.1 & 14.6 & 9.0 & 89.2 & 16.3 & 23.2 \\
        Skywork-PRM-7B* & 61.8 & 82.9 & 70.8 & 43.8 & 62.2 & 53.6 & 17.9 & 31.9 & 22.9 & 14.0 & 41.9 & 21.0 & 42.1 \\
        % EurusPRM-Stage1 & 46.9 & 42.0 & 44.3 & 33.3 & 38.2 & 35.6 & 23.9 & 19.8 & 21.7 & 21.9 & 24.5 & 23.1 & 31.2 \\
        % EurusPRM-Stage2 & 51.2 & 44.0 & 47.3 & 36.4 & 35.0 & 35.7 & 25.7 & 18.0 & 21.2 & 23.1 & 19.1 & 20.9 & 31.3 \\
        ReasonEval-7B & 26.1 & 95.3 & 41.0 & 35.7 & 77.6 & 48.9 & 27.5 & 55.2 & 36.7 & 27.0 & 60.6 & 37.4 & 41.0 \\
        Qwen2.5-Math-7B-PRM800K* & 53.1 & 95.3 & 68.2 & 48.0 & 90.1 & 62.6 & 35.7 & \textbf{87.3} & 50.7 & 29.8 & \textbf{86.1} & 44.3 & 58.5 \\
        Qwen2.5-Math-PRM-7B* & \textbf{72.0} & 96.4 & \textbf{82.4} & 68.0 & \textbf{90.4} & \textbf{77.6} & 55.7 & 85.5 & \textbf{67.5} & 55.2 & 83.0 & \textbf{66.3} & \textbf{73.5} \\
        % Dis-PRM-7B & 65.7 & 88.1 & 75.2 & 68.2 & 75.6 & 71.7 & 55.5 & 57.8 & 56.6 & 54.2 & 55.6 & 54.9 & 64.6 \\
        \text{$\bigstar$} R-PRM-7B-SFT & 66.2 & 92.7 & 77.2 & 60.3 & 88.2 & 71.6 & 48.6 & 77.3 & 59.6 & 40.1 & 75.5 & 52.3 & 65.2 \\
        \text{$\bigstar$} R-PRM-7B-DPO & \textbf{72.0} & 91.7 & 80.7 & \textbf{71.2} & 83.5 & 76.9 & \textbf{60.2} & 67.8 & 63.8 & \textbf{55.5} & 65.6 & 60.1 & 70.4 \\
        % \noalign{\vspace{4pt}}
        % \hline
        % \noalign{\vspace{4pt}}
        % % \hline
        % \multicolumn{14}{l}{\textbf{72B}} \\
        % Qwen2.5-Math-RM-72B & 41.1 & 46.1 & 43.5 & 39.7 & 58.1 & 47.2 & 28.1 & 56.6 & 37.6 & 18.8 & 50.2 & 27.4 & 38.9 \\
        % \text{$\bigstar$} Qwen2.5-Math-PRM-72B & 78.7 & 97.9 & \textbf{87.3} & 74.2 & 88.2 & \textbf{80.6} & 67.9 & 82.0 & \textbf{74.3} & 64.8 & 78.8 & \textbf{71.1} & \textbf{78.3} \\
        \bottomrule[1.5pt]
    \end{tabular}
}
\caption{Performance on ProcessBench. \text{$\bigstar$} represents the models we trained. Results marked with * come from \citeauthor{qwenprm} Bold text denotes the best results within PRM.  }
\label{tab:processbench}
\end{table*}
\begin{table*}[!t]
\centering
\scriptsize
\renewcommand{\arraystretch}{1.2}
\resizebox{\textwidth}{!}{%
    %\begin{tabular}{l |c|ccc| ccccc| cc cc}
    \begin{tabular}{l ccc ccccc cc ccc}
    \toprule[1.5pt]
    \multirow{2}{*}{\textbf{Model Name}}  & \multicolumn{3}{c}{\textbf{Simplicity}}  & \multicolumn{5}{c}{\textbf{Soundness}}& \multicolumn{4}{c}{\textbf{Sensitivity}}& \multirow{2}{*}{\textbf{Overall}}\\
    \cmidrule(lr){2-4} \cmidrule(lr){5-9} \cmidrule(lr){10-13} 
    % \cline{3-5} \cline{6-10} \cline{11-14}
    & \textbf{NR.} & \textbf{NCL.} & \textbf{Avg.} &\textbf{ES} &\textbf{SC.}&\textbf{DC.} &\textbf{CI} & \textbf{Avg.} &\textbf{PS} & \textbf{DR.} & \textbf{MS.} & \textbf{Avg.}   \\
    % Random & 50.0 & 50.0& 50.0& 50.0& 50.0& 50.0& 50.0& 50.0& 50.0& 50.0& 50.0& 50.0& 50.0\\
    \midrule
    \multicolumn{14}{l}{\textit{\textbf{LLM-as-judge, Proprietary language models}}} \\
    %\midrule \multicolumn{14}{@{}c@{}}{\textit{\textbf{LLM-as-judge, Proprietary language models}}} \\ \midrule
    GPT-4o*  & 57.0 & 62.4 & 59.7 & 72.0 & 69.7 & 70.7 & 71.1 & 70.9 & 62.5 & 65.7 & 99.2 & 75.8 & 66.8\\
    o1-mini*  & 65.6 & 63.7 & 64.6 & 74.5 & 67.7 & 73.8 & 72.3 & 72.1 & 61.8 & 64.8 & 100.0 & 75.5 & 68.8\\
    \midrule
    \multicolumn{14}{l}{\textit{\textbf{PRMs}}} \\
    Math-Shepherd-7B*  & 44.0 & 50.3 & 47.1 & 49.4 & 44.5 & 41.3 & 47.7 & 45.7 & 47.2 & 48.6 & 86.1 & 60.7 & 47.0\\
    Math-PSA-7B  & 47.6 & 55.1 & 51.3 & 56.5 & 49.4 & 47.1 & 54.2 & 51.8 & 51.7 & 54.1 & 88.9 & 64.9 & 52.3\\
    RLHFlow-Mistral-8B*  & 46.1 & 47.3 & 46.7 & 56.6 & 55.1 & 54.4 & 63.8 & 57.5 & 51.5 & 56.2 & 97.9 & 68.5 & 54.4\\
    RLHFlow-DeepSeek-8B*  & 46.4 & 48.9 & 47.6 & 55.7 & 55.0 & 53.2 & 66.2 & 57.5 & 49.0 & 55.4 & \textbf{99.8} & 68.1 & 54.2\\
    Llemma-PRM800k-7B*  & 49.3 & 53.4 & 51.4 & 56.4 & 47.1 & 46.7 & 53.3 & 50.9 & 51.0 & 53.5 & 93.6 & 66.0 & 52.0\\
    %Llemma-MetaMath-7B*  & 48.7 & 49.3 & 49.0 & 54.2 & 46.8 & 44.5 & 53.5 & 49.8 & 49.2 & 51.3 & 91.8 & 64.1 & 50.3\\
    Skywork-PRM-7B*  & 35.7 & 41.2 & 38.4 & 36.7 & 29.1 & 30.6 & 34.4 & 32.7 & 36.8 & 37.4 & 88.8 & 54.3 & 36.2\\
    ReasonEval-7B*  & \textbf{61.0} & 50.1 & 55.5 & 62.1 & 65.9 & 61.5 & 66.0 & 63.9 & 55.6 & 58.0 & 99.5 & 71.0 & 60.0\\
    %ReasonEval-34B*  & 54.8 & 48.1 & 51.5 & 66.4 & 60.3 & 57.8 & 67.5 & 63.0 & 57.7 & 64.3 & 97.2 & 73.1 & 60.5\\
    Qwen2.5-Math-7B-PRM800K  & 48.6 & 47.8 & 48.2 & 62.1 & 59.4 & 58.7 & 68.5 & 62.2 & 52.9 & 64.0 & \textbf{99.8} & 72.2 & 58.3\\
    Qwen2.5-Math-PRM-7B  & 49.0 & 55.1 & 52.1 & 71.8 & 67.3 & 66.3 & \textbf{78.5} & 71.0 & 57.6 & 69.1 & 99.7 & 75.5 & 65.5\\
    %Dis-PRM-7B  & 52.1 & \textbf{64.6} & \textbf{58.4} & 66.3 & 58.5 & 58.4 & 63.5 & 61.7 & 59.0 & 62.9 & 98.0 & 73.3 & 61.4 \\
     \text{$\bigstar$} R-PRM-7B-SFT & 52.7 & \textbf{64.7} & \textbf{58.7} & 70.1 & 62.7 & 63.4 & 69.5 & 66.4 & \textbf{61.4} & 67.4 & 98.3 & 75.7 & 64.9 \\
    \text{$\bigstar$} R-PRM-7B-DPO  & 52.2 & 58.2 & 55.2 & \textbf{72.1} & \textbf{69.1} & \textbf{68.9} & 75.0 & \textbf{71.2} & 61.2 & \textbf{69.5} & 99.1 & \textbf{76.6} & \textbf{66.8}\\ 
    % Gemini-2.0-flash-exp & 66.0 & 67.2 & 58.1 & 62.7 & 70.4 & 65.7 & 66.0 & 67.3 & 67.3 & 61.8 & 66.2 & 98.2 & 75.4\\
    % Gemini-2.0-thinking-exp-1219 & 68.8 & 68.5 & 63.8 & 66.2 & 72.9 & 71.3 & 71.0 & 71.8 & 71.8 & 60.3 & 65.7 & 99.8 & 75.3\\
    \bottomrule[1.5pt]

    \end{tabular}
}
\caption{Performance on PRMBench. \text{$\bigstar$} represents the models we trained. Results marked with * come from \citeauthor{qwenprm} Bold text denotes the best results within PRM. }
\label{tab:prmbench}
\vspace{-\baselineskip} % 可选：进一步压缩下方间距
\end{table*}

\subsection{Experiment Settings}

\paragraph{Tasks and Benchmarks:} To validate the effectiveness of our method for process-level reward modeling, we evaluate it on two challenging benchmarks: ProcessBench\cite{processbench} and PRMBench\cite{prmbench}.

\begin{itemize}[noitemsep, topsep=0pt]
    \item \textbf{ProcessBench}~\cite{processbench} is designed to assess the ability to identify erroneous steps in mathematical reasoning processes. The dataset comprises 3,400 test cases covering mathematical problems of varying difficulty levels. Each test case includes a question accompanied by an LLM generated step-by-step solution, with the earliest incorrect step meticulously annotated by human experts. The primary task for the model is to either accurately identify the erroneous step or confirm the correctness of all steps in the solution.
    
    % \item \textbf{PRMBench}~\cite{prmbench} is a benchmark specifically designed to evaluate PRMs, covering fine-grained error detection capabilities during inference. It contains 6,216 problems with 83,456 step-level annotations, designed to evaluate process-level reward models' capabilities in detecting various types of reasoning errors across multiple dimensions including simplicity, soundness, and sensitivity.
    %This benchmark dataset contains 6,216 complex reasoning problems accompanied by 83,456 meticulously annotated reasoning steps. 
   \item \textbf{PRMBench}~\cite{prmbench} constitutes a comprehensive benchmark for evaluating PRMs, with particular emphasis on granular error diagnosis. It assesses PRM capabilities across three key dimensions: Simplicity, Soundness, and Sensitivity. These dimensions are further subdivided into nine specific aspects. A detailed description of these nine aspects is provided in Appendix~\ref{sec:prmench-description}.
   %Through its multi-dimensional evaluation framework encompassing logical coherence, computational parsimony, and error sensitivity metrics, PRMBench enables systematic assessment of PRMs' capacity to identify and classify diverse reasoning errors at different procedural stages. 

% : Non-Redundancy (NR), Non-Circular Logic (NCL), Empirical Soundness (ES), Step Consistency (SC), Domain Consistency (DC), Confidence Invariance (CI), Prerequisite Sensitivity (PS), Deception Resistance (DR), and Multi-Solution Consistency (MS).

% This greedy search strategy ensures that each step is the locally optimal choice, aiming to improve the quality and consistency of the generated content. However, since it always selects the highest-scoring step at each stage, it may overlook the globally optimal solution. As a result, in certain complex tasks, combining this approach with other search strategies may lead to better overall performance.

% This benchmark consists of 6,216 problems and 83,456 stepwise annotations, covering three major evaluation criteria: simplicity, reliability, and sensitivity. Furthermore, it categorizes errors into nine distinct types, including redundant logic, domain inconsistency, and deceptive reasoning. By assessing model performance across different error types, \textit{PRMBench} provides researchers with a comprehensive tool for diagnosing model weaknesses and guiding improvements, thereby facilitating the advancement and application of process-level reward models in multi-step reasoning tasks.
\end{itemize}
Furthermore, we validate the effectiveness of the reward model by applying it to guide the test-time scaling of the policy model Best-of-N, Greedy Guide Search, separately. Specifically, we evaluate on multiple benchmarks, including \textit{MATH500}~\cite{verify-step-by-step}, \textit{Minerva Math}~\cite{minerva_math}, \textit{OlympiadBench}~\cite{olympiadbench}, \textit{College Math}~\cite{college_math}, \textit{AIME24}, \textit{AMC23}~\footnote{Due to the large size of the Olympiad and College Math test sets, we randomly selected 200 samples from each to form the test subsets.}. Following \citet{qwenprm}, we use Qwen2.5-7B-Instruct to sample eight candidate steps with a temperature of 1.

 \begin{itemize}[noitemsep, topsep=0pt]
    \item \textbf{Best-of-N:} selects the response with the highest score among $N$ candidates, as evaluated by a PRM. 

    \item \textbf{Greedy Guide Search:} at each generation step, the model generates $N$ candidate continuations and selects the one with the highest score, as determined by a PRM, to extend the reasoning trajectory. This procedure is repeated until a complete solution is produced.
\end{itemize}

\paragraph{Baselines:} We selected the following strong process reward models as baselines.
\begin{itemize}[noitemsep, topsep=0pt]
    \item \textbf{Math-Shepherd}~\cite{Math-shepherd}: Automatically obtaining the probability of reaching the correct solution as step labels based on Monte Carlo Tree Search (MCTS).

    \item \textbf{Math-PSA}~\cite{openr}: combining existing automatic annotation techniques~\cite{omegaprm} and integrating data from Math-Shepherd and PRM800K datasets.
    
    \item \textbf{RLHFlow-DeepSeek/Mistral}~\cite{rlhflow}: Similar to Math-Shepherd, but trained with iterative DPO.
    %A process reward model (PRM) developed based on Online Iterative Reinforcement Learning from Human Feedback (RLHF), aiming to improve alignment in large language models (LLMs). Unlike traditional offline RLHF methods, RLHFlow incorporates on-policy sampling and proxy preference modeling to iteratively refine reward signals.

    % A series of process reward models (PRM) developed from Qwen-2.5-Math-Instruct, similar to MATH-Shepherd, using the probability of reaching the correct answer as a label, but additionally incorporates LLM-as-Judge for consistency filtering.

    %Unlike PRM800K, which relies on costly human annotations, MATH-APS uses OmegaPRM~\cite{omegaprm} for efficient automated data collection by identifying and correcting errors in reasoning paths.
    \item \textbf{Skywork-PRM-7B}~\cite{skyworkprm}: based on Qwen2.5-Math-Instruct and recently released by Skywork.

    %A process reward model (PRM) developed based on Qwen-2.5-Math, designed to enhance reward modeling efficiency through data-centric optimization. Unlike conventional PRMs relying on large-scale datasets, Skywork-PRM employs high-quality, filtered preference data (Skywork-Reward Preference 80K) to improve label accuracy and model robustness.
    \item \textbf{ReasonEval-7B}~\cite{reasoneval}: Evaluates mathematical problem-solving step by step, assessing validity and redundancy.
    %A process reward model (PRM) designed to evaluate the quality of reasoning steps in mathematical problem-solving, going beyond final-answer accuracy. It introduces validity and redundancy metrics to assess whether each step is both correct and necessary for problem-solving.

    \item \textbf{Llemma-PRM800K-7B}~\cite{llemma}: Trained exclusively on PRM800K from levels 1 through 3.

    \item \textbf{Qwen2.5-Math-PRM-7B}~\cite{qwenprm}: Based on Math-Shepherd, additionally employs LLM-as-Judge to perform consistency filtering and result in a 1.8M samples training dataset.

% \SSJ{baseline need rewrite here}

    % A process reward model (PRM) designed for easy-to-hard generalization, enabling reward models trained on easier mathematical problems (Level 1-3) to evaluate and guide reasoning on harder problems (Level 4-5) without direct human supervision.

\end{itemize}

\paragraph{Implement details:} We process each PRM800K case by prompting LLaMA3.3-70B-Instruct to generate four candidate evaluation trajectories, resulting in approximately 289k SFT samples and 269k DPO samples. We train the model on Qwen2.5-Math-7B-Instruct using one epoch of fine-tuning, with a batch size of 128 and separate learning rates: 5e-6 for SFT and 5e-7 for DPO. We set aside 20k samples for validation and retain the checkpoint with the lowest validation loss. By default, we report results using ten evaluation reasoning trajectories per step.

\subsection{Experiment Results}
\begin{table*}[htp]
\centering
% \resizebox{1\textwidth}{!}
\footnotesize
{
\begin{tabular}{@{}lccccccc@{}}
\toprule[1.5pt]
\textbf{Setting} & \textbf{AIME24} & \textbf{AMC23} & \textbf{\makecell{MATH}} & \textbf{\makecell{Olympiad\\Bench}} & \textbf{\makecell{College\\MATH}} & \textbf{\makecell{Minerva\\MATH}} & \textbf{Avg.} \\ \midrule
pass@1 & 11.2 & 47.8 & 73.0 & 38.0 & 38.6 & 37.2 & 41.0 \\
%greedy           & 10.0 & 52.5 & 77.4 & 39.0 & 40.5 & 41.9 & 43.6 \\ 
major@8          & 20.0 & 57.5 & 79.6 & 47.0 & 41.5 & 42.7 & 48.0 \\
pass@8(Upper Bound)           & 33.3 & 82.5 & 88.8 & 58.5 & 47.5 & 57.7 & 61.4 \\ 
\midrule
\textbf{Greedy Guide Search@8} & & & & & & &  \\
Math-Shepherd-7B    & 13.3 & 52.5 & 74.6 & 38.5 & 36.5 & 41.2 & 42.8 \\
Math-PSA-7B        & 6.7  & 57.5 & 79.8 & 42.5 & \textbf{41.0} & 39.3 & 44.5 \\
RLHFlow-PRM-Mistral-8B & 10.0 & 57.5 & 73.4 & 37.5 & 38.0 & 41.2 & 42.9 \\
RLHFlow-PRM-DeepSeek-8B & 13.3 & 52.5 & 74.8 & 39.5 & 37.0 & 40.8 & 43.0 \\
Llemma-PRM800K-7B & 13.3 & 57.5 & 73.8 & 40.0 & 36.5 & 38.2 & 43.2 \\
%Llemma-MetaMath-7B & 6.7  & 42.5 & 73.2 & 40.0 & 37.0 & 39.0 & 39.7 \\
Skywork-PRM-7B   & 10.0 & 57.5 & 77.8 & 41.5 & 39.0 & 43.4 & 44.9 \\
ReasonEval-7B    & 3.3  & 55.0 & 73.0 & 37.5 & 35.5 & 37.9 & 40.4 \\
Qwen2.5-Math-7B-PRM800K & \textbf{23.3} & 45.0 & 78.2 & 42.0 & 35.5 & 38.6 & 43.8 \\
Qwen2.5-Math-PRM-7B & 16.7 & 60.0 & \textbf{81.0} & 43.5 & 39.0 & 40.4 & 46.8 \\
\text{$\bigstar$} \method-7B-DPO & 16.7 & \textbf{70.0} & 80.0 & \textbf{46.5} & 39.5 & \textbf{43.4} & \textbf{49.4} \\
\bottomrule[1.5pt]
\end{tabular}
}
\caption{The performance of PRM guided greedy search  with policy model
Qwen2.5-7B-Instruct.}
\vspace{-\baselineskip} % 可选：进一步压缩下方间距
\label{tab:guide_search}
\end{table*}

\paragraph{\method achieves high evaluation accuracy efficiently.} As shown in Table~\ref{tab:processbench} and Table~\ref{tab:prmbench},  our SFT approach achieves F1 scores of 65.2 and 64.9 on ProcessBench and PRMBench, respectively, representing significant improvements over state-of-the-art baselines. Notably, when compared with Qwen2.5-Math-7B-PRM800K (the strongest PRM800K-based baseline), \method achieves F1 improvements of 6.7 and 6.6 points on ProcessBench and PRMBench, respectively. The effectiveness of our framework is further demonstrated by its 7.2-point and 3.3-point F1 improvements over LLaMA3.3-70B-Instruct and GPT-4o, respectively, on ProcessBench. These results demonstrate that our reasoning-driven evaluation paradigm successfully enhances model learning efficiency while maximizing the utility of human-annotated data.

The model's capabilities are significantly  enhanced through DPO, achieving remarkable performance scores of 70.4 and 66.8 on two benchmarks. Notably, it surpassed Qwen2.5-Math-PRM (trained with 1.8M data points) on the PRMbench evaluation. These experimental results conclusively validate the efficacy of the DPO approach, highlighting the potential of generative evaluation paradigms to substantially improve assessment capabilities.
\vspace{0.5em}

\paragraph{\method provides more comprehensive evaluation across multiple dimensions.} In rigorous benchmarking with PRMBench, \method-DPO demonstrates significant advantages over Qwen2.5-Math-7B-PRM800K. It achieves improvements of 7.0, 9.0, and 4.4 points across the three core evaluation dimensions. Notably, it surpasses GPT-4 in both completeness and sensitivity, positioning itself as a more comprehensive assessment paradigm.

More specifically, \method's strength in soundness is reflected in three dimensions: empirical validity, step consistency, and domain alignment.
 This structural strength directly enhances the model's ability to detect logical errors through progressive contextual analysis. Each reasoning step undergoes a rigorous examination of preceding steps to identify inconsistencies. Notably, \method exhibits strong prerequisite sensitivity, with a particular advantage in deception resistance over o1-mini. This highlights its robustness in detecting reasoning steps that appear superficially valid but are logically flawed—errors that conventional systems often fail to identify.
\vspace{0.5em}

\begin{table*}[htp]
\centering
\footnotesize
% \resizebox{1\textwidth}{!}
{
\begin{tabular}{@{}lcccccccc@{}}
\toprule[1.5pt]
\textbf{Setting} & \textbf{AIME24} & \textbf{AMC23} & \textbf{\makecell{MATH}} & \textbf{\makecell{Olympiad\\Bench}} & \textbf{\makecell{College\\Math}} & \textbf{\makecell{Minerva\\MATH}} & \textbf{Avg.} \\ \midrule
pass@1 & 11.2 & 47.8 & 73.0 & 38.0 & 38.6 & 37.2 & 41.0 \\
maj@8 & 20.0 & 57.5 & 79.6 & 47.0 & 41.5 & 42.7 & 48.0 \\
pass@8 & 33.3 & 82.5 & 88.8 & 58.5 & 47.5 & 57.7 & 61.4 \\
 \midrule
\textbf{\textsc{7B+}} & & & & & & & \\
Math-Shepherd-7B & 16.7 & 42.5 & 76.0 & 42.0 & 37.0 & 39.3 & 42.3 \\
Math-PSA-7B & \textbf{20.0} & 55.0 & 80.8 & 47.5 & 39.5 & 40.1 & 47.2 \\
RLHFlow-Mistral-8B & 10.0 & 55.0 & 76.8 & 42.0 & 39.5 & 37.1 & 43.4 \\
RLHFlow-DeepSeek-8B & 13.3 & 57.5 & 76.2 & 40.0 & 39.0 & 39.7 & 44.3 \\
Llemma-PRM800K-7B & 10.0 & 52.5 & 76.6 & 42.5 & 39.0 & 42.7 & 43.9 \\
%Llemma-MetaMath-7B & \textbf{20.0} & 50.0 & 76.6 & 40.5 & 40.5 & 40.4 & 44.7 \\
Skywork-PRM-7B & 16.7 & 55.0 & 81.2 & 44.0 & 40.5 & \textbf{44.5} & 47.0 \\
ReasonEval-7B & 6.7 & 55.0 & 75.2 & 41.0 & 40.0 & 40.4 & 43.1 \\
Qwen2.5-Math-7B-PRM800K & 13.3 & 57.5 & 80.0 & 44.5 & \textbf{43.5} & 43.0 & 47.7 \\
Qwen2.5-Math-PRM-7B & 16.7 & 55.0 & 82.0 & 48.0 & \textbf{43.5} & 43.0 & \textbf{48.0} \\ 
\text{$\bigstar$} R-PRM-7B-DPO & \textbf{20.0} & \textbf{62.5} & \textbf{82.2} & \textbf{48.0} & 41.0 & 44.1 & \textbf{49.6} \\
\bottomrule[1.5pt]
\end{tabular}
}
\caption{Performance comparison on the Best-of-8 strategy of the policy model Qwen2.5-7B-Instruct.
\text{$\bigstar$} represents the models we trained.}
\label{tab:best-of-n}
\end{table*}

\paragraph{\method demonstrates superior generalization capability.}   As shown in Table~\ref{tab:processbench}, all listed open-source PRMs, except Qwen2.5-Math-PRM-7B and Skywork-PRM-7B for which the training data sources are unknown, have been trained exclusively on GSM8K and MATH. Among these PRMs, only Math-PSA-7B and Qwen2.5-Math-7B-PRM800K achieve F1 scores above 60 on certain ProcessBench subsets, while others perform relatively poorly, particularly on out-of-domain datasets such as OmniMATH and OlympiadBench. Except for Qwen2.5-Math-7B-PRM800K, which achieved an F1 score of 50.7 on OlympiadBench, the remaining models scored below 50, with most falling between 10 and 40. By contrast, \method not only performs well on the MATH dataset but also achieves F1 scores above 60 on all out-of-domain datasets. This suggests that \method has learned a dataset-independent reasoning pattern, enabling it to perform well across datasets with varying difficulty.

\vspace{0.5em}

\paragraph{\method guides policy model to reach correct answer effectively.} As shown in Table~\ref{tab:guide_search} and Table~\ref{tab:best-of-n}, our method achieves 8.4\% and 8.6\% average accuracy improvements over the Pass@1 baseline in the Guide Search and Best-of-N settings, respectively. It further achieves state-of-the-art performance by outperforming Qwen2.5-MATH-PRM by 2.6 and 1.6 points, and surpassing Majority Voting in both settings. The experimental results directly demonstrate that our method's accurate reward evaluation at each reasoning step effectively guides the policy model to arrive at correct solutions. Furthermore, we believe our approach holds greater potential for integration with backtracking-enabled strategies like Monte Carlo Tree Search (MCTS) and multi-candidate strategies such as Beam Search, enabling more comprehensive utilization of our methodology.
% PRMBench evaluates PRM capabilities across three key dimensions: simplicity, soundness, and sensitivity, providing a more thorough understanding of PRM’s strengths and weaknesses. The evaluation results of PRMBench are presented in Table \ref{tab:prmbench}. Apart from non-redundancy, where \method-DPO underperforms compared to ReasonEval-7B\cite{reasoneval}, it surpasses PRMs of the same scale in all other aspects. Notably, ReasonEval is specifically trained to assess the correctness and redundancy of steps, whereas \method has not undergone targeted training for this aspect.

% \input{latex/tables/data_source}

\section{Analysis}
\subsection{Effective Data Scaling}
\begin{figure}[htbp]
    \centering
    \includegraphics[width=0.5\textwidth]{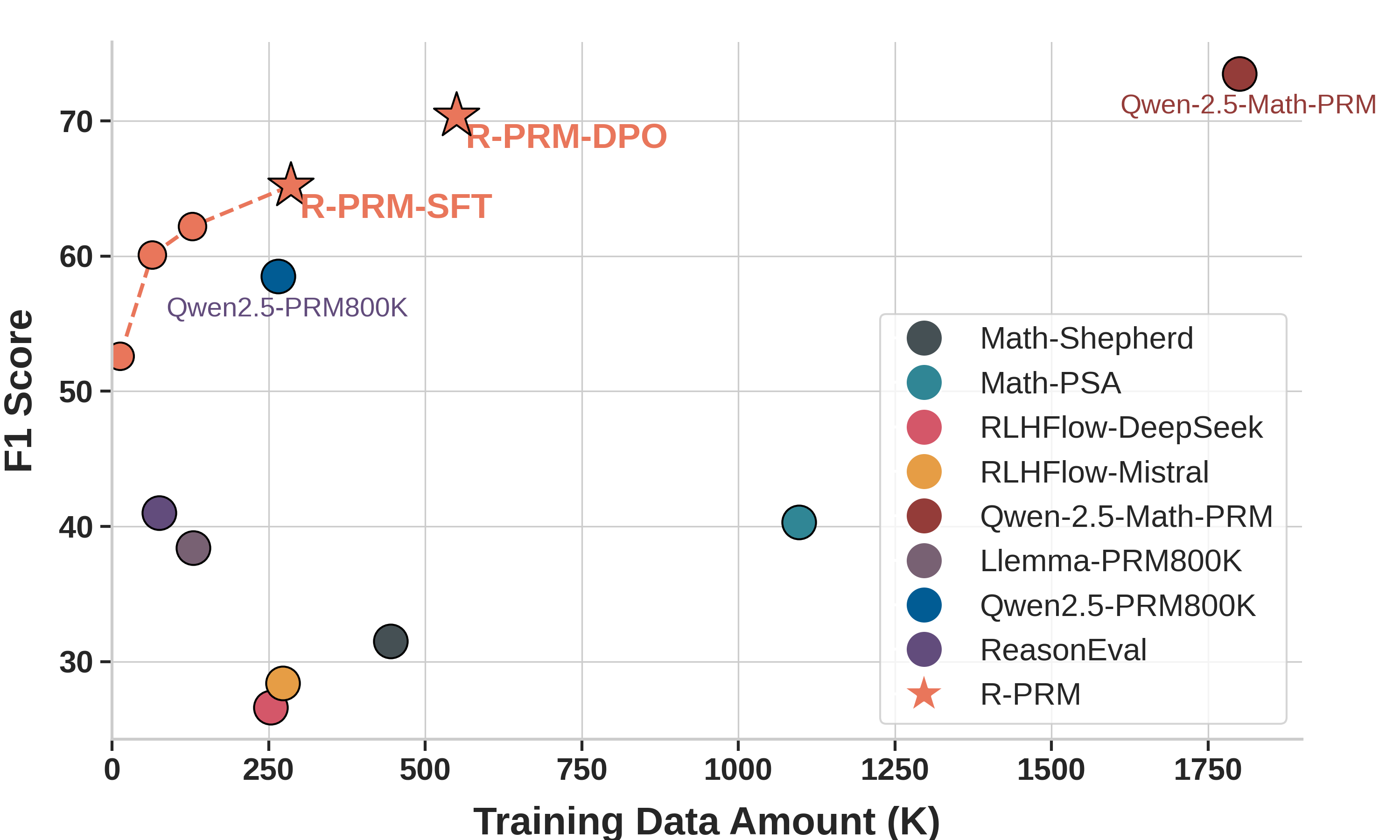}
    \caption{Average F1 score on ProcessBench with different training data scales.}
    \label{fig:data-scaling}
    \vspace{-\baselineskip} % 可选：进一步压缩下方间距
\end{figure}

We conduct comprehensive experiments to study the effect of training data scale on model performance. Figure~\ref{fig:data-scaling} illustrates the performance variation of \method on ProcessBench under varying training set sizes. With only 12.8k training samples, \method achieves an impressive F1 score of 52.6, which already surpasses most existing open-source PRMs. When scaled up to 64k samples, \method demonstrates superior performance, outperforming Qwen2.5-Math-7B-PRM800K, which is trained on 265k samples, by a margin of 1.6 points. Further scaling the training data to the full set of 285k samples yields substantial improvements, reaching an F1 score of 65.2.

Without requiring additional labeled data, we leverage 269K preference pairs from existing sampling results, effectively improving the model’s performance from 65.2 to 70.4 and demonstrating exceptional efficiency in data utilization.

\subsection{Inference-Time-Scaling}
\begin{figure}[htbp]
    \centering
    \includegraphics[width=0.5\textwidth]{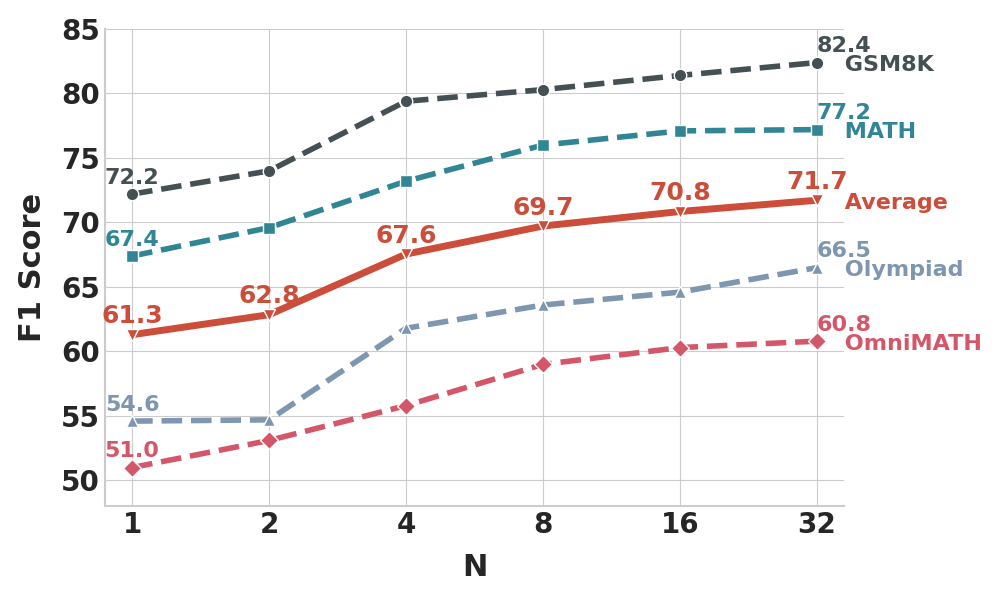}
    \caption{Efficient scaling inference-time compute on ProcessBench.}
    \vspace{-\baselineskip} % 可选：进一步压缩下方间距
    \label{fig:processbench-scaling}
\end{figure}

We conducted an investigation into how \method's performance scales with increasing inference-time budgets. As shown in Figure \ref{fig:processbench-scaling}, \method demonstrates consistent performance gains on ProcessBench as the number of evaluation trajectories increases. Notably, scaling from 2 to 4 samples leads to a substantial F1 improvement from 62.8 to 67.6 on ProcessBench. Moreover, increasing the number of evaluation trajectories consistently yields performance improvements across all four datasets, which demonstrates the robustness of our scaling strategy and highlights a unique advantage of our reasoning-driven approach.

% We further investigated how \method performance varies when given additional inference budgets. As shown in Figure \ref{fig:processbench-scaling}, the performance of \method on ProcessBench increases with the number of evaluation trajectories sampled, particularly when the sample size grows from 2 to 4, where the ProcessBench F1 score rises from 62.8 to 67.6. Specifically, when the number of CoT samples increased from 1 to 32, the performance improvements of 10.2, 9.8, 9.8, and 11.9 were observed across the four datasets, further demonstrating the effectiveness of test-time scaling. In contrast, other PRMs are unable to leverage this capability.

% \vspace{-1em}
\subsection{Threshold Robustness in Model Evaluation}
\begin{figure*}[t]
    \centering
    \includegraphics[width=0.90\textwidth]{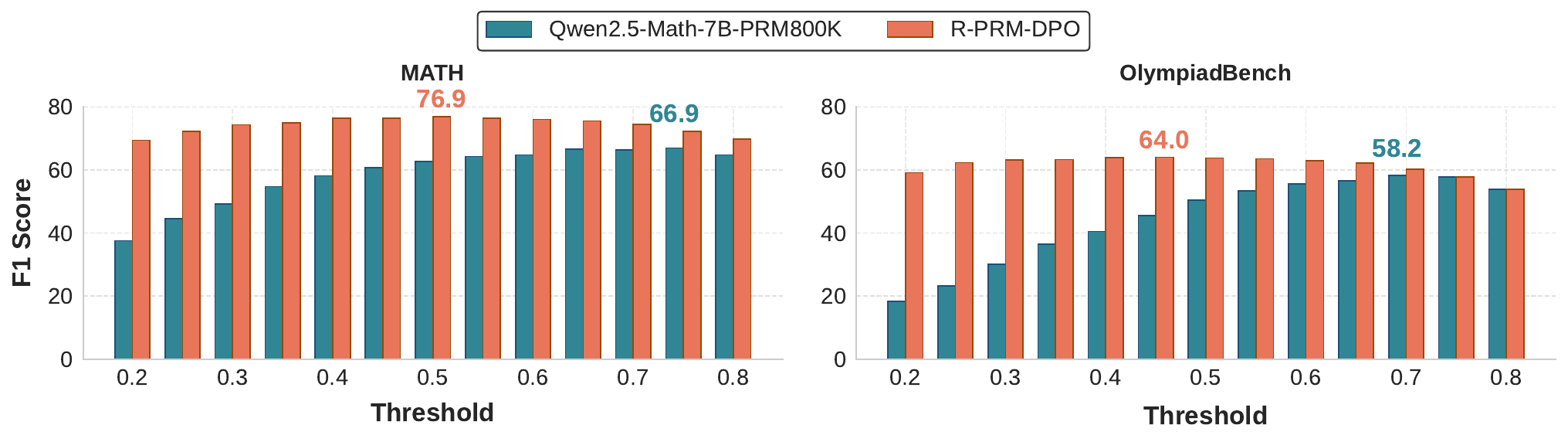}
    \caption{Average F1 score on ProcessBench at different thresholds, with the overall performance on the left and performance on the OlympiadBench dataset on the right.
}
    \label{fig:processbench-different-threshold}
\end{figure*}

\begin{figure*}[t]
\vspace{-0.2cm}
\footnotesize
\centering
\begin{tabular}{ |p{\dimexpr\linewidth-4\tabcolsep-1.5pt}| }
\toprule

\multicolumn{1}{|p{\dimexpr\linewidth-4\tabcolsep-1.5pt}|}{
{\textbf{Problem}: An $\textit{emirp}$ is an integer that, when its digits are written in reverse order, is a prime number. For example, 73 is an emirp because 37 is prime. What is the smallest two-digit prime number that is NOT an emirp?}
} \\
\midrule
{\textbf{Solution}: \ldots \newline
%Step1: To determine the smallest two-digit prime number that is not an emirp, we need to follow these steps: First,\ \ldots\ldots\newline 
Step7: Continue to the next smallest two-digit prime number. 19 is prime.\newline
{\color{red}Step 8: Check if 23 is an emirp. Reverse the digits of 23: \(32\). Check if 32 is prime: No, 32 is not prime (divisible by 2). Since 23 is a prime number and its reverse, 32, is not a prime number, 23 is not an emirp. \ldots}
} \\
\midrule
{\textbf{Qwen2.5\_Math-PRM-7B} correctness score: {\color{red}0.9883}}
\\
\midrule
{\textbf{Qwen2.5\_Math-7B-PRM800K} correctness score: {\color{red}0.8594}}
\\
\midrule
{\textbf{R-PRM}: Analysis: Previous Steps Analysis:  \ \ldots\newline
%Now Step Analysis: \ \ldots\ldots
\ldots\ldots\newline
Calculation Analysis:  \ \ldots\ldots  \textbf{\color{blue}Upon reviewing, the mistake in the analysis is assuming 23 is the answer without properly checking if there are smaller primes that are not emirps}. \ \ldots\ldots \textbf{\color{blue}Given the oversight in the analysis regarding the thorough checking of primes and their reversals, the conclusion that 23 is the smallest such prime without properly validating against all primes in sequence is premature.} \ \ldots\ldots\newline
Conclusion: \textbf{\color{blue}The Now Step incorrectly concludes that 23 is the smallest two-digit prime number \ \ldots\ldots\newline} \
Verification: {\color{blue}Is the step correct (Yes/No)? No}
} \\
\midrule
{\textbf{R-PRM (Majority Voting)} score: {\color{blue}0.0547}}
\\
\bottomrule
\end{tabular}
\caption{
 A case study from ProcessBench MATH dataset. {\color{red}Red} text denotes the error step and the scores of other models, and the {\color{blue}blue} text highlights our model's critique of the error and our score for that step.
}
\vspace{-0.3cm}
\label{case-study-3}
\end{figure*}

During evaluations of ProcessBench and PRMBench, we adopt a fixed threshold of 0.5 for binary classification to determine whether each step is correct. We further analyze the model's sensitivity to threshold variations. As shown in Figure~\ref{fig:processbench-different-threshold}, \method demonstrates strong robustness to threshold variation, with minimal performance fluctuations on ProcessBench. In contrast, Qwen2.5-Math-7B-PRM800K exhibits greater sensitivity, showing a noticeable rightward shift in its performance curve on ProcessBench.

On the more challenging out-of-domain test subset OlympiadBench, our method maintains its robustness to threshold variations, while Qwen2.5-Math-7B-PRM800K exhibits a larger threshold shift, indicating its tendency to misclassify incorrect solutions as correct. This demonstrates that our method has better generalization capability, maintaining more accurate evaluation even when problem types and domains change. For detailed performance analysis of scores and thresholds on PRMBench, please refer to the Appendix \ref{sec:threshold-on-prmbench}.

% On the OlympiadBench dataset, as shown on the right side of Figure \ref{fig:processbench-different-threshold}, \method continues to show strong robustness without a clear threshold preference. In contrast, Qwen2.5-Math-7B-PRM800K again shows a rightward shift in threshold bias. Despite both models being trained on PRM800K, \method outperforms Qwen2.5-Math-7B-PRM800K in terms of robustness and stability. For detailed performance analysis of scores and thresholds on PRMBench, please refer to the Appendix \ref{sec:threshold-robustness-on-prmbench}.
%As illustrated in Figure \ref{fig:prmbench-different-threshold}, the experimental results of PRMBench show that \method has significant robustness advantages, while Qwen2.5-Math-7B-PRM800K exhibits a performance gap of 8.2 points between the 0.5 threshold and its optimal performance.

\subsection{Case Study}

As shown in Figure~\ref{case-study-3}, the solution erroneously skipped verifying the number 19 in Step 7 and directly proceeded to check number 23 in Step 8.  Unfortunately, both strong baselines, Qwen2.5-Math-PRM and Qwen2.5-Math-7B-PRM800K, failed to detect the omission, mistakenly assigning high reward scores to Step 8 (0.99 and 0.86, respectively). In contrast, \method first analyzed both the previous and current steps. Based on this analysis, \method concluded that the task required verifying the numbers in ascending order, which showcases its advanced logical reasoning capabilities. Subsequently, \method resumed the reasoning process for Step 7 to verify the correctness of number 19, thus identifying the discrepancy between its own result and the answer in Step 8. Through this reasoning process, \method assigned a reward score of 0.05 to Step 8, successfully detecting the error. Please refer to the Appendix \ref{sec:additional-case} for more cases.

\section{Conclusion}
In this paper, we present Reasoning-Driven Process Reward Modeling (R-PRM), a novel framework that advances the process reward modeling of mathematical reasoning. Our framework consists of three components. First, we leverage stronger LLMs to construct seed data, enabling our model to perform a comprehensive evaluation process. Second, we use preference optimization to enhance performance without requiring additional annotated data. Third, we introduce inference-time scaling to fully harness the model's reasoning capabilities. Extensive experiments demonstrate that our method achieves significant performance improvements on ProcessBench and PRMBench, while also effectively guiding LLM reasoning. Further analysis shows that R-PRM exhibits more comprehensive, robust, and generalizable evaluation capabilities, as its performance continues to improve with increased inference, highlighting its substantial potential.

\section*{Limitations}
Due to computational resource constraints, we have not yet verified our approach on larger models such as 70B, despite extensive experiments demonstrating its effectiveness on 7B models. We hypothesize that larger models, given their enhanced reasoning capabilities, could achieve higher modeling accuracy when combined with our methodology. Additionally, while we have tested popular inference strategies like Best-of-N and Guided Search, our exploration of advanced search algorithms remains limited. Sophisticated methods such as Monte Carlo Tree Search (MCTS) and Beam Search remain underexplored, although they could potentially better leverage the characteristics of PRM and yield improved generation results.

\bibliography{custom}

\appendix

% \section{Example Appendix}
% \label{sec:appendix}
\section{Detailed Description of PRMBench Subcategories}
\label{sec:prmench-description}
\begin{itemize}
    \item \textbf{Non-Redundancy (NR)}: Evaluates the model's ability to identify and eliminate unnecessary steps within the reasoning process, ensuring efficiency without sacrificing correctness.
    
    \item \textbf{Non-Circular Logic (NCL)}: Assesses whether the model can detect circular reasoning, where conclusions are reintroduced as premises, leading to logical loops.
    
    \item \textbf{Empirical Soundness (ES)}: Measures the model's capability to identify and reject reasoning steps that contradict established facts or real-world knowledge.
    
    \item \textbf{Step Consistency (SC)}: Evaluates whether the reasoning steps maintain consistency with each other, ensuring that all steps logically flow from one to the next.
    
    \item \textbf{Domain Consistency (DC)}: Assesses the model's ability to apply domain-specific knowledge correctly, avoiding the misuse of concepts or theories across different domains.
    
    \item \textbf{Confidence Invariance (CI)}: Tests whether the model maintains appropriate confidence levels throughout the reasoning process, especially when errors are detected or uncertainties arise.
    
    \item \textbf{Prerequisite Sensitivity (PS)}: Evaluates whether the model detects missing prerequisites or conditions essential for valid reasoning, ensuring the completeness of the logic.
    
    \item \textbf{Deception Resistance (DR)}: Measures the model's ability to detect and reject misleading information that might appear correct but contains subtle errors.
    
    \item \textbf{Multi-Solution Consistency (MS)}: Assesses the model's ability to handle multiple valid solutions to the same problem, ensuring consistency across different reasoning paths.
\end{itemize}

\section{Threshold Robustness on PRMBench}
\label{sec:threshold-on-prmbench}
\begin{figure}[htbp]
    \centering
    \includegraphics[width=0.5\textwidth]{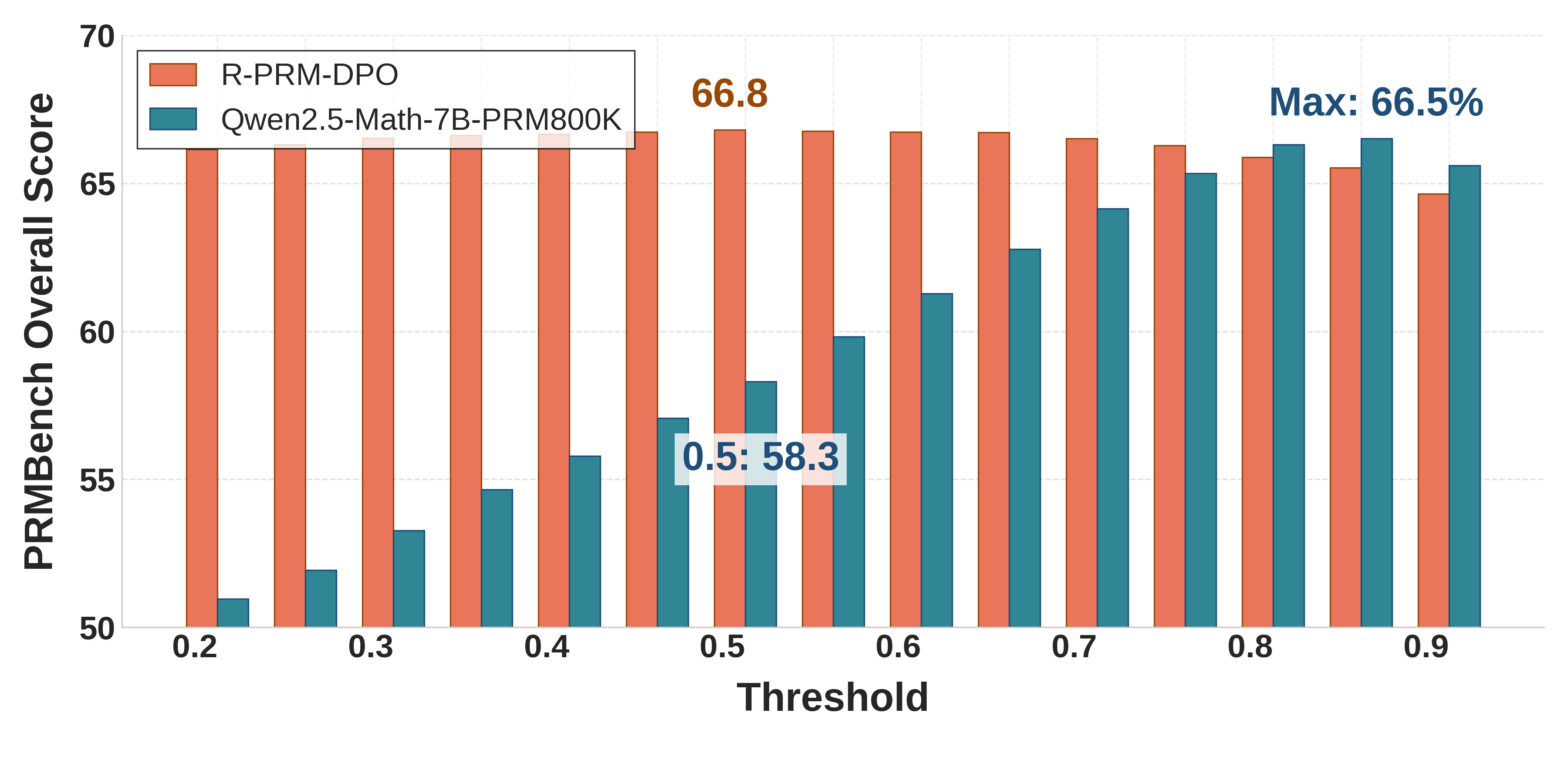}
    \caption{PRMBench Overall Score with Different Threshold}
    \label{fig:prmbench-different-threshold}
\end{figure}

As illustrated in Figure \ref{fig:prmbench-different-threshold}, the experimental results of PRMBench show that \method has significant robustness advantages, while Qwen2.5-Math-7B-PRM800K exhibits a performance gap of 8.2 points between the 0.5 threshold and its optimal performance.
\label{sec:threshold-robustness-on-prmbench}

\section{Additional Case}
As shown in this case, both Qwen2.5-Math-PRM-7B and Qwen2.5-Math-7B-PRM800K fail to detect the calculation error in this step, yet they show high confidence in their judgments with their correctness scores around 0.9. On the other hand, our R-PRM shows its strong ability by explaining the step, checking the calculation, and finding the error, with a final score of around 0.3.
\label{sec:additional-case}
\begin{figure*}[t]
\vspace{-0.2cm}
\footnotesize
\centering
\begin{tabular}{ |p{\dimexpr\linewidth-4\tabcolsep-1.5pt}| }
\toprule

\multicolumn{1}{|p{\dimexpr\linewidth-4\tabcolsep-1.5pt}|}{
{\textbf{Problem}: Suppose $t$ is a positive integer such that $\mathop{\text{lcm}}[12,t]^3=(12t)^2$. What is the smallest possible value for $t$?}
} \\
\midrule
{\textbf{Solution}: 

To solve this problem, we can start by using the properties of the least common multiple (LCM). Given: $\mathop{\text{lcm}}[12,t]^3=(12t)^2$.

{\color{red}Step 1: Take the cube root of both sides of the equation. $\sqrt[3]{\mathop{\text{lcm}}[12,t]^3}=\sqrt[3]{(12t)^2}$. This simplifies to: $\mathop{\text{lcm}}[12,t]=12t^{2/3}$.}

% Step 2: Since the LCM of two numbers is always a multiple of both numbers, we can express $\mathop{\text{lcm}}[12,t]$ as $12k$ where k is an integer. So we have: $12k=12t^{2/3}$
\ldots
Step 31: We can see that $13824=2304^2$, so $t=48$ satisfies the original equation. Therefore, the smallest possible value for $t$ is $\boxed{48}$.
} \\
\midrule
{\textbf{Qwen2.5-Math-PRM-7B} score: {\color{red}0.9414}}
\\
\midrule
{\textbf{Qwen2.5-Math-7B-PRM800K} score: {\color{red}0.8711}}
\\
\midrule
{\textbf{R-PRM (Ours)}: Analysis: Previous Steps Analysis: Step 1: This step states the problem and sets up the equation to be solved, which is $\mathop{\text{lcm}}[12,t]^3=(12t)^2$.This step is a correct restatement of the problem\ \ldots\ldots\newline
\ldots \newline
Calculation Analysis: The calculation involves taking the cube root of $(12t)^2$. By the properties of exponents, $\sqrt[3]{(12t)^2} = \sqrt[3]{12^2 \cdot t^2} = 12^{2/3} \cdot t^{2/3}$. This is because the cube root of a product can be expressed as the product of the cube roots, and the cube root of a power can be expressed as that power divided by 3. {\color{blue}Therefore, the correct simplification should be $\mathop{\text{lcm}}[12,t] = 12^{2/3} \cdot t^{2/3}$, not $12t^{2/3}$ as stated in the Now Step.}\newline
Conclusion: {\color{blue}The Now Step incorrectly simplifies the expression $\sqrt[3]{(12t)^2}$ to $12t^{2/3}$. The correct simplification should be $12^{2/3} \cdot t^{2/3}$, which is different from what is presented. Therefore, the Now Step contains an error.}\newline
Verification: {\color{blue}Is the step correct (Yes/No)? No"}
} \\
\midrule
{\textbf{R-PRM} score: {\color{blue}0.2992}}
\\
\bottomrule
\end{tabular}
\caption{
A case study from MATH subset of ProcessBench. {\color{red}Red} text denotes the error step and the scores of other models, and the {\color{blue}blue} text highlights our model's critique of the error and our score for that step.}
\vspace{-0.3cm}
\label{cot-verifier-example-main-3}
\end{figure*}

\section{Used Scientific Artifacts}
Below are the scientific artifacts used in our work.
For the sake of ethics, our use of these artifacts is consistent with their intended use.
\begin{itemize} [itemsep=1pt]
    \item \textit{Transformers (Apache-2.0 license)}, a framework to facilitate downloading and training state-of-the-art pretrained models.
    \item \textit{trl (Apache-2.0 license)}, a full stack library that provides a set of tools to train transformer language models with Reinforcement Learning. The library is built on top of the Transformers library.
    
    \item \textit{PRM800K(MIT license)}, a large-scale dataset proposed by OpenAI, containing 800,000 step-level correctness labels for model-generated solutions to problems from the MATH dataset.
    
    \item  \textit{PRMBench(Apache-2.0 license)}, a benchmark designed to evaluate the ability of PRMs to identify error-types in the solution process, consisting of 6,216 fine-grained data instances.
\end{itemize}

\section{Prompt for Construction Data}
\label{sec:construction-data-prompt}
\onecolumn
\begin{table}[t]
    \centering
    \small
    \begin{xltabular}{\textwidth}{>{\raggedright\arraybackslash}X}
    \toprule
You are an excellent math teacher. Please verify the correctness of the Now Step.\newline

You first need to analyze the Now Step and the Previous Steps and then summarize based on your analysis.\\
Analysis:\\
You need to analyze the following aspects.\\
**Previous Steps Analysis**: You need to analyze the Previous Steps step by step. For each step, you need to first explain what the current step is doing, then you try to find any error in the current step.\\
**Now Step Analysis**: You first need to explain what the Now Step is doing, and then point out which part of the Question it is trying to solve or which part of the information it states.\\
**Data Source Analysis**: First you need to find out what data are used in the Now Step, and then you need to determine whether the source of the data is reasonable and correct. When you judge whether the source of a data is reasonable and correct, you need to specify the specific source of this data: such as which part of the question, or which content of the previous step; and then determine the source and current use is consistent, the Now Step is used correctly.\\
**Consistency Analysis**: You need to check that the Now Step is consistent with the contents of the Previous Steps, and then you need to check that all the information inside the Now Step is consistent.\\
**Calculation Analysis**: If the Now Step involves any calculations, such as addition, subtraction, multiplication, division, equations, modulo operations, etc., you will first need to perform a check on the calculation, such as a reverse operation, to see if the calculation was done correctly, and then analyze the results of your check to see if there was an error in the calculation.\\
Conclusion: \\
Please verify the correctness of the Now Step based on your analysis, if there is any error in the Now Step then the Now Step is wrong and vice versa the Now Step is correct. At the end of the Conclusion, when you give your final answer, write it in the form "Verification: Is the step correct (Yes/No)? X", where X is either Yes or No.\newline

Question: [Math Problem]\\
Previous Steps: [Previous Steps]\\
Now Step: [Current Step]\\
Please carefully analyze the correctness of the Now Step.\\
Reply:  \\
     \bottomrule
     \end{xltabular}
    \caption{The Prompt to Construct Data }
    \label{tab:r0_template}
\end{table}
\twocolumn

\end{document}